\documentclass[letterpaper]{article}

\usepackage[authoryear]{natbib}
\usepackage{alifeconf}
\usepackage{multirow}
\usepackage{graphicx}
\usepackage{epstopdf}

\title{Quantifying the Impact of Parameter Tuning on Nature-Inspired Algorithms}
\author{Matthew Crossley, Andy Nisbet \and Martyn Amos\\
\mbox{} \\
School of Computing, Mathematics and Digital Technology,\\
Manchester Metropolitan University, Manchester M1 5GD, UK. \\
Email: m.crossley@mmu.ac.uk}

\begin{document}
\maketitle

\begin{abstract}
The problem of {\it parameterization} is often central to the effective deployment of nature-inspired algorithms. However, finding the optimal set of parameter values for a combination of problem instance and solution method is highly challenging, and few concrete guidelines exist on how and when such {\it tuning} may be performed. Previous work tends to either focus on a specific algorithm or use benchmark problems, and both of these restrictions limit the applicability of any findings. Here, we examine a {\it number} of different algorithms, and study them in a ``problem agnostic" fashion (i.e., one that is not tied to specific instances) by considering their performance on  {\it fitness landscapes} with varying characteristics. Using this approach, we make a number of observations on which algorithms may (or may not) benefit from tuning, and in which specific circumstances. 
\end{abstract}

\section{Introduction and Background}

There exist many algorithms that are inspired by nature, and each has associated with it a set of {\it parameters}. These define specific features or details of an algorithm that may be altered in order to change the behaviour or performance of the method (for example, in evolutionary algorithms, parameters may include mutation rate or crossover probability).  The problem  of finding the optimal settings for these parameters (often referred to as ``tuning") is well-established \citep{lobo2007parameter,nannen2008costs,beetuning,birattari2009tuning,eiben2011parameter}, but little in-depth work has been performed on quantifying the {\it benefits} of tuning for a {\it range} of algorithms. We address this in the current paper, by investigating the precise benefits (or otherwise) of tuning for a number of different algorithms. Moreover, we do this in a way that is independent of any specific {\it problem}, by using an approach based on fitness landscape characteristics. The main contribution of the paper is therefore to establish a framework for deciding - prior to any problem-specific implementation - which algorithms may (or may {\it not}) benefit from tuning. Our aim is to offer advice to future practitioners on the relative merits of tuning, compared to the effort involved in finding the best set of parameter values. We achieve this by establishing, for each algorithm, the problem features that offer the most {\it potential} for performance improvements via tuning.

Previous work  \citep{crossley2013} characterised a number of nature-inspired algorithms according to their performance on fitness landscapes with different features. However, the authors used the {\it default} parameter settings for each algorithm, which fails to reflect the fact that, in practice, methods are usually {\it tuned} prior to serious use  \citep{leung2003tuning,adenso2006fine,koster2007importance,ridge2010tuning}.  Here, we extend this work by quantifying the relative merits of tuning for a range of algorithms in a wide variety of fitness landscape scenarios. We achieve this by assessing both their tuned and untuned behaviour, using the methods described in \citet{crossley2013}.

In order to tune our selected algorithms, we use the notion of {\it racing}, which was first introduced in the field of machine learning \citep{maron1993hoeffding, Maron97theracing}. Specifically, we use the F-race algorithm \citep{birattari2002racing,yuan2004statistical,smit2009comparing,birattari2010f}, which has been extensively used to find the best possible set of parameter values for a given problem in a limited time.

The rest of the paper is organised as follows: we first describe our approach in the Methodology section, before presenting our experimental findings in the Results section. We conclude with a discussion of the implications of our results, and suggest further work.

\section{Methodology}

Our methodology may be summarised as follows: (1) select a number of nature-inspired algorithms, and obtain consistent source code for their implementation; (2) for each algorithm, find the best parameter settings (i.e., tune) over a number of different problems; (3) compare the performance of tuned and untuned algorithms.

\subsection{Algorithm selection}

We compare a number of nature-inspired algorithms, all of which are commonly applied to continuous function optimisation (we use the same set as in \citet{crossley2013}). These may be classified \citep{Brabazon2006} as either {\it social}, {\it evolutionary} or {\it physical}. The social algorithms we select are Bacterial Foraging Optimisation Algorithm (BFOA) \citep{Passino2002}, Bees Algorithm (BA) \citep{Pham2006}, and Particle Swarm Optimisation (PSO) \citep{Kennedy1995}. The evolutionary algorithms selected are Genetic Algorithms (GA) \citep{Goldberg1989} and Evolution Strategies (ES) \citep{Back1993}, and physical algorithms are represented by Harmony Search (HS) \citep{Geem2001}. We also include stochastic hill climbing (SHC) as a ``baseline" algorithm; in contrast to \citet{crossley2013} we exclude random search, as it has no parameters to tune. As before, we heed the observation that ``Ideally, competing algorithms would be coded by the same expert programmer and run on the same test problems on the same computer configuration"  \citep{Barr1995}. With that in mind, we use only implementations provided by \citet{Brownlee2011}.  Space prevents a complete description of specific implementation details for each algorithm, but full implementation details can be found in  \citet{Brownlee2011}, which is freely available and contains complete source code.

\subsection{Tuning}

Our fundamental goal is to investigate the pre- and post-tuned performance of our selected algorithms on landscapes with different general features, and thus identify characteristics of landscapes for which tuning may yield significant differences in algorithm performance. As  \citet{Morgan2011} observe, ``Different problem types have their own characteristics, however it is usually the case that complementary insights into algorithm behaviour result from conducting larger experimental studies using a variety of different problem {\it types}" (our emphasis).  Rather than using arbitrary benchmark instances of problems in order to perform tuning, we use a landscape-based approach, as utilised in \citet{crossley2013}. As \citet{Morgan2011} explain, this Max-Set of Gaussians (MSG) method \citep{Gallagher2006} is a ``randomised landscape generator that specifies test problems as a weighted sum of Gaussian functions. By specifying the number of Gaussians and the mean and covariance parameters for each component, a variety of test landscape instances can be generated. The topological properties of the landscapes are intuitively related to (and vary smoothly with) the parameters of the generator." We now describe the characteristics under study:

{\it Ruggedness} of a landscape is often linked to its difficulty \citep{Jones1995}, and factors affecting this include (1) the {\it number} of local optima  \citep{Horn1994}, and (2) {\it ratio} of the fitness value of local optima to the global optimal value \citep{Malan2009,Freisleben2000}.  Other significant factors concern (3) {\it dimensionality} \citep{Hendtlass2009} (that is, the number of variables in the objective function), (4) {\it boundary constraints} (that is, the limits imposed on the value of a variable) \citep{Kukkonen}, and (5) {\it smoothness} of each Gaussian curve (effectively the gradient) used to generate the landscape \citep{Beyer2002} - a smaller value indicates a smoother gradient. For each characteristic, we use the same ranges as in \citet{crossley2013}, summarised in Table \ref{featurespace}.

		\begin{table}[!t]
			\begin{center}
				\caption[]{A summary of the ranges selected for the various characteristics in the landscape generation methodology.}
				\label{featurespace}
				\begin{tabular}[c]{p{3cm} c c c c }
					\hline\noalign{\smallskip}
					{\bf Characteristic} & {\bf Min} & {\bf Max} & {\bf Step} & {\bf Default} \\
					\noalign{\smallskip}
					\hline
					\noalign{\smallskip}
					No. of local optima & 0 & 9 & 1 & 3\\
					Ratio of local optima to global optimum & 0.1 & 0.9 & 0.2 & 0.5\\
					Dimensionality & 1 & 10 & 1 & 2\\
					Boundary constraints & 10 & 100 & 10 & 30\\
					Smoothness & 10 & 100 & 10 & 15\\
					\hline
				\end{tabular}
			\end{center}
		\end{table}

To produce a test set of problems, we use the MSG landscape generator.  For every value of every characteristic (in the range specified in Table \ref{featurespace}) we generate a set of five landscapes, which makes up the initial problem set for each value. We then use the F-racing methodology \citep{birattari2002racing} to find optimised parameters for each algorithm, over {\it every} value of {\it every} landscape characteristic used.  We ensure that termination criteria are standardised, in order to ensure reasonable comparisons, and therefore use the number of objective function evaluations to determine when to terminate an algorithm's run. We established, through initial experiments, that all selected algorithms generally converge within 20,000 objective function evaluations, so we use that as the specific value.

\subsection{Comparison}	

We run each algorithm 100 times on each landscape in the set of landscapes generated for each particular characteristic value (when investigating smoothness, for example, we generate 1000 different landscapes (100 each for smoothness = 10 \dots 100), and run each algorithm 100 times on each landscape). This is done first for all algorithms with `default' parameter configurations, and then again, this time using the parameter configurations obtained through the F-Racing process. We measure the performance of each algorithm in terms of the mean ($\mu$) and standard deviation ($\sigma$) of the exact average error obtained, over all values for a particular characteristic. That is, we investigate the {\it robustness} of each algorithm to changes in the values for each characteristic, rather than their absolute performance on specific problem instances. This allows us to identify specific landscape features where tuning may make a significant difference, some difference, or no difference at all, for a particular algorithm.

\section{Results}

We find that the effect of tuning using F-Racing is varied across algorithms, and that they fit into three categories: Algorithms which {\it do not} benefit from F-Racing (ES), algorithms which only benefit significantly from F-Racing when a landscape is `difficult' for the algorithm using default parameters (BA, HS, PSO), and algorithms which {\it always} benefit from F-Racing (BFOA, GA, SHC).  Of course, we acknowledge the fact that F-Racing is just one of many possible meta-search techniques available for parameter tuning, and future work will involve a comparative study of alternative methods.

We summarise our results in Table \ref{sumresults}; the full datasets are available online\footnote{http://dx.doi.org/10.6084/m9.figshare.696908}; the repository contains all performance data across all runs, summary spreadsheets and details of all parameter settings. We now examine in detail the performance of each algorithm, using spider plots to graphically depict the results in Table \ref{sumresults}. For each plot, the further a line is from the origin, the {\it smaller} the average error (that is, the ``larger" an area, the larger the degree of robustness, which is considered ``better").

	\begin{table*}
		\caption{Mean ($\mu$) and standard deviation ($\sigma$) of the exact average error of algorithm performance (both untuned (UT) and tuned (T)). Smaller values imply more robustness to changes in a specific characteristic.}
		\scriptsize
		\centering
		\begin{tabular}[!t]{p{1.2cm} | c || c | c || c | c || c | c || c | c || c | c || c | c || c | c }
		 \multicolumn{2}{c||}{ } & \multicolumn{2}{|c||}{BFOA} & \multicolumn{2}{|c||}{Bees Algorithm} & \multicolumn{2}{|c||}{ES}& \multicolumn{2}{|c||}{GA}& \multicolumn{2}{|c||}{Harmony Search}& \multicolumn{2}{|c||}{PSO}& \multicolumn{2}{|c}{SHC}\\
		\hline
		\multicolumn{2}{c||}{ } & UT & T & UT & T &UT & T &UT & T &UT & T &UT & T &UT & T  \\
		\hline
\multirow{2}{1.3cm}{\# of Local Optima}  &  $\mu$ & 0.118 & 0.003 & 0.001 & 8.8$\times10^{{}-6}$ & 0.085 & 0.078 & 0.093 & 0.015 & 0.011 & 2.2$\times10^{-5}$ & 0.025 & 0.014 & 0.266 & 0.072 \\
 &  $\sigma$ & 0.011 & 0.001 & 2.1$\times10^{-4}$ & 9.2$\times10^{-7}$ & 0.028 & 0.026 & 0.033 & 0.008 & 0.005 & 2.9$\times10^{-5}$ & 0.010 & 0.010 & 0.041 & 0.020 \\
\hline
\multirow{2}{1.3cm}{Dimensions}  &  $\mu$ & 0.754 & 0.417 & 0.216 & 0.073 & 0.542 & 0.544 & 0.420 & 0.529 & 0.364 & 0.263 & 0.420 & 0.157 & 0.577 & 0.589 \\
 &  $\sigma$ & 0.388 & 0.360 & 0.202 & 0.069 & 0.345 & 0.346 & 0.233 & 0.401 & 0.271 & 0.204 & 0.307 & 0.145 & 0.261 & 0.371 \\
\hline
\multirow{2}{1.4cm}{Local Optima Ratio}  &  $\mu$ & 0.120 & 0.003 & 0.001 & 8.7$\times10^{-5}$ & 0.084 & 0.082 & 0.079 & 0.007 & 0.007 & 0.002 & 0.025 & 0.016 & 0.284 & 0.088 \\
 &  $\sigma$ & 0.021 & 0.002 & 2.3$\times10^{-4}$ & 1.9$\times10^{-4}$ & 0.012 & 0.012 & 0.006 & 0.006 & 0.003 & 0.004 & 0.004 & 0.011 & 0.045 & 0.011 \\
\hline
\multirow{2}{1.3cm}{Boundary Range}  &  $\mu$ & 0.317 & 0.022 & 0.001 & 0.001 & 0.097 & 0.093 & 0.125 & 0.021 & 0.048 & 0.001 & 0.076 & 0.022 & 0.446 & 0.305 \\
 &  $\sigma$ & 0.213 & 0.033 & 1.3$\times10^{-4}$ & 0.001 & 0.017 & 0.018 & 0.057 & 0.016 & 0.041 & 0.001 & 0.050 & 0.013 & 0.239 & 0.217 \\
\hline
\multirow{2}{1.3cm}{Smoothness}  &  $\mu$ & 0.260 & 0.010 & 0.004 & 0.001 & 0.110 & 0.102 & 0.154 & 0.021 & 0.018 & 0.001 & 0.043 & 0.014 & 0.349 & 0.112 \\
 &  $\sigma$ & 0.089 & 0.005 & 0.002 & 4.3$\times10^{-4}$ & 0.012 & 0.012 & 0.045 & 0.011 & 0.007 & 0.001 & 0.012 & 0.006 & 0.039 & 0.014 \\
		\end{tabular}
		\label{sumresults}
	\end{table*}
	
	\subsection{Bacterial Foraging Optimisation Algorithm}

\begin{figure}[t]
		\centering
		\includegraphics[width=0.4\textwidth]{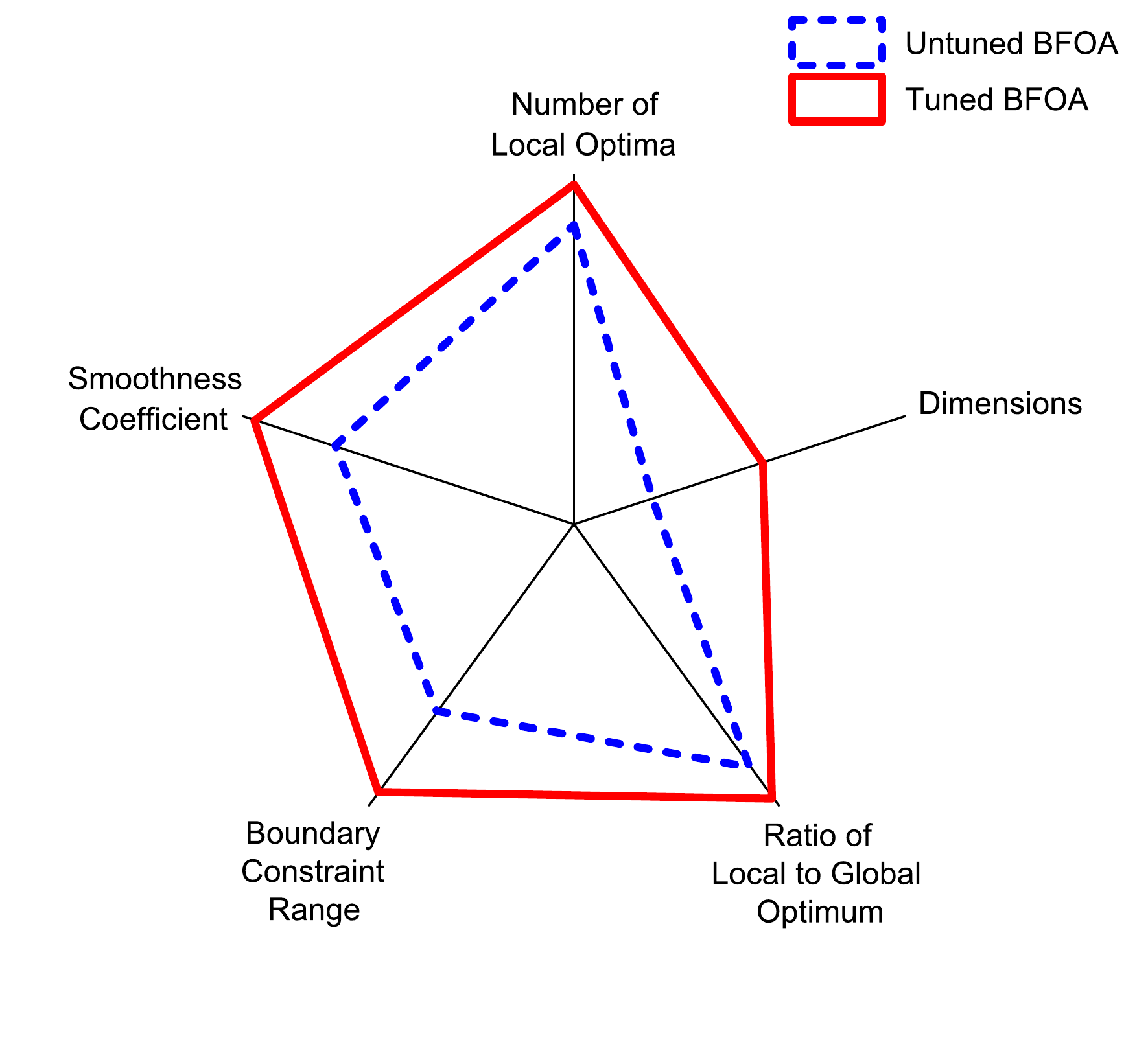}
		\caption{Summary of results for  {\bf Bacterial Foraging Optimisation Algorithm}.}
		\label{bfoasp}
	\end{figure}

There exists little discussion on the role of different parameters in the BFOA.  While some elements of the search pattern are clearly altered by various parameters, it is very difficult to estimate values for these.  In the original description of the BFOA \citep{Passino2002}, the parameter values were assigned based on observation of actual bacterial colonies.  While this may be true to the nature-inspired concept, it is not necessarily the best way to obtain optimal performance from the algorithm.
The combination of parameters offered by BFOA gives a highly configurable search environment.  Parameters such as step size and population size directly affect the potential area the algorithm can explore in a given number of objective function calculations.  Attraction and repulsion weights, and the ``space'' over which these attraction and repulsion effects spread, work to control local optima avoidance.  Parameters controlling the number of chemotactic steps before a reproduction step, and the number of reproduction steps before an elimination-dispersal event, control the balance of {\it exploitation} versus {\it exploration}.   Given that the search behaviour of the algorithm is highly configurable, it is unsurprising that BFOA is heavily reliant on tuning.  Results for BFOA are shown in Figure \ref{bfoasp}. Across all characteristics, tuning offers a {\it significant} improvement on the average error and standard deviation of the performance - in many cases, improving the ranking of the algorithm from the largest average error to one of the smallest, and coping well with the changing characteristics.   We see the most significant improvement where boundary constraint ranges change, a characteristic that is heavily reliant on parameters which control the range at which new solutions are generated (in the case of BFOA, this is the {\it step size}).  Improvements are also shown for dimensionality and smoothness coefficient, increasing the performance of BFOA where there is little gradient information in a large fitness landscape.  Smaller improvements are demonstrated by the increasing number of local optima and the increasing attractiveness of these local optima, but tuning still benefits the algorithm considerably.

In terms of the configurations selected by F-Racing, there is little variation in parameter values as characteristics change.  Across all characteristics, and all values for those characteristics, there are only eight different configurations selected by racing.  This suggests that, while it is difficult to find a {\it good} configuration, once it has been found, it is likely to be good for all {\it similar} problems.  Tuning is vital to the performance of the BFOA, but it is possible that by exploring problems using a similar methodology to that demonstrated here, we may create a `bank' of promising configurations.

\subsection{Bees Algorithm}

\begin{figure}[t]
		\centering
		\includegraphics[width=0.4\textwidth]{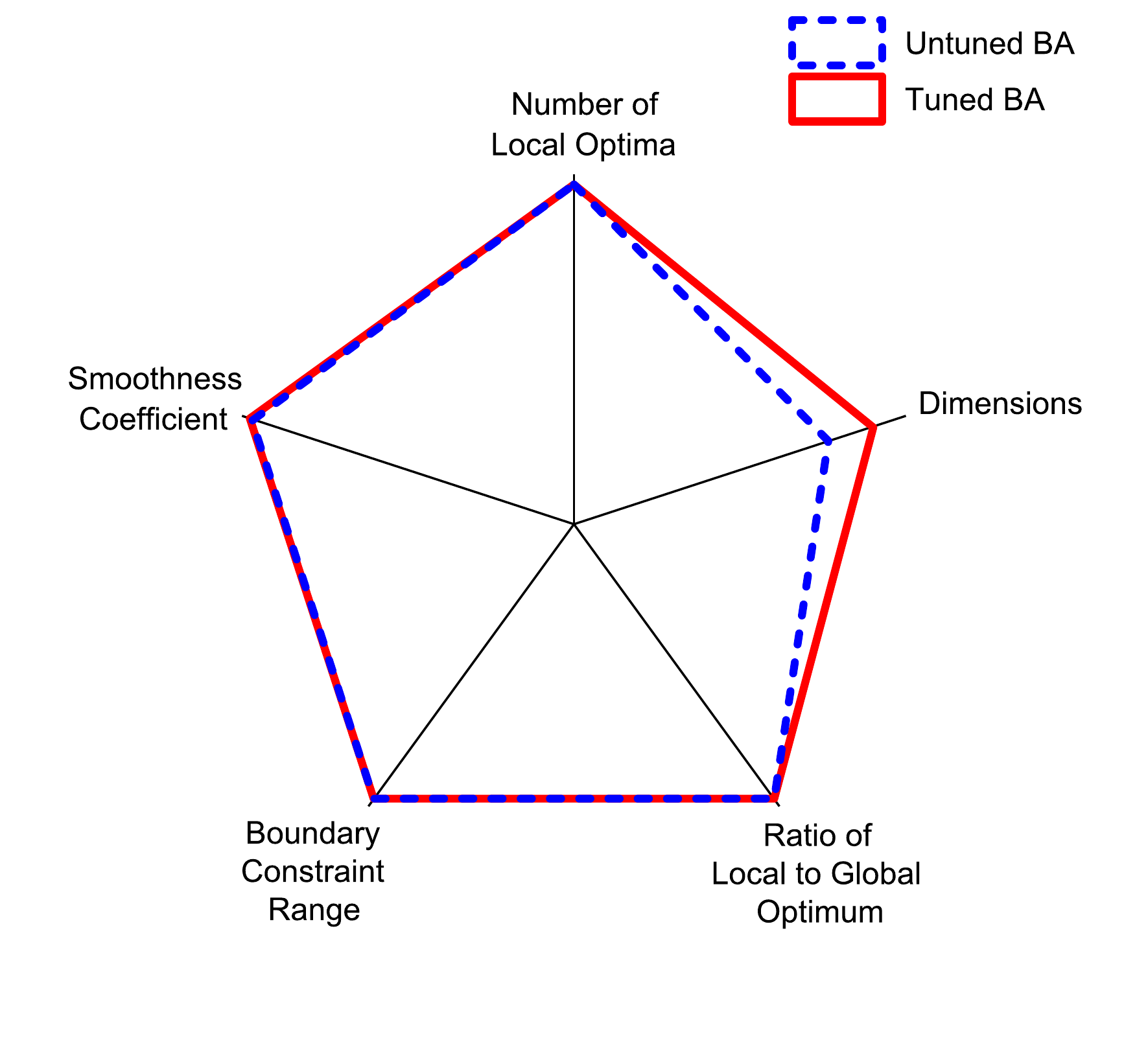}
		\caption{Summary of results for  {\bf Bees Algorithm}.}
		\label{beesp}
	\end{figure}

The BA is considered to be an algorithm on which parameterisation has little effect \citep{Pham2006}.   We observe that the BA is one of the best {\it untuned} performers in this study, offering weight to this argument for relative parameter insensitivity. In terms of adjusting the BA to cope with an increasing number of local optima, there are several parameters which have an effect.  Parameters such as the {\it number of sites} under investigation, the {\it number of bees} attributed to those sites, and the {\it differentiation} between sites and `elite' sites are all factors which affect the searching behaviour of the algorithm to allow for greater flexibility as the modality of the problem landscape increases.   Results for BA are shown in Figure \ref{beesp}.  Post-tuning, we find that the BA selects the {\it same} parameter configuration, regardless of the number of local optima present in the landscape.  We then see that tuning has no effect on the ability of the algorithm to cope with increasing numbers of local optima.  As long as the number of sites under investigation is greater than the number of optima, the algorithm is capable of dealing with modality.  Coupled with the abandonment of `unpromising' sites, this means that `too many' sites are not detrimental to the exploration pattern of the algorithm.  

We see the same pattern when increasing the ratio of local optima to the global optimum.  As long as the number of sites under investigation covers the modality of the landscape, the BA is not hampered by increasing levels of attractiveness, regardless of parameter settings.  The {\it patch size} parameter of the BA controls the distance from a site bees are allowed to explore.  This is the parameter which affects the search behaviour of the algorithm as boundary constraint size increases.  The BA allows for full coverage of any sized search space, using {\it scout bees} to investigate new random sites to give `teleportation' across the landscape.  As with the number of local optima, we find the F-Races for the BA select the same parameter set for most boundary constraint sizes.  We find that, post-tuning, the performance of the BA actually {\it decreases} slightly, suggesting the algorithm can cope less well with changes in boundary constraint size.  We believe that the configurations may have become over-fitted to the landscapes used for tuning, and, while performance on the landscapes used for racing may have increased, the ability to search generalised landscapes has decreased. Dimensionality provides the most significant result in terms of pre-tuning and post-tuning performance of the BA.  We observe little change in performance at one to three dimensions - the point where the untuned algorithm is already performing well.  As dimensionality increases beyond this, the effect of tuning becomes {\it increasingly} beneficial. We suggest that there is no increase in performance in other characteristics because these landscapes are {\it not challenging enough} to the BA to require adjusting the parameters.  For the ranges of landscape characteristics on which we have tested the BA, it is clear that tuning generally makes little difference to the performance, as suggested by its original developer.  

\subsection{Evolution Strategies}

	\begin{figure}[t]
		\centering
		\includegraphics[width=0.4\textwidth]{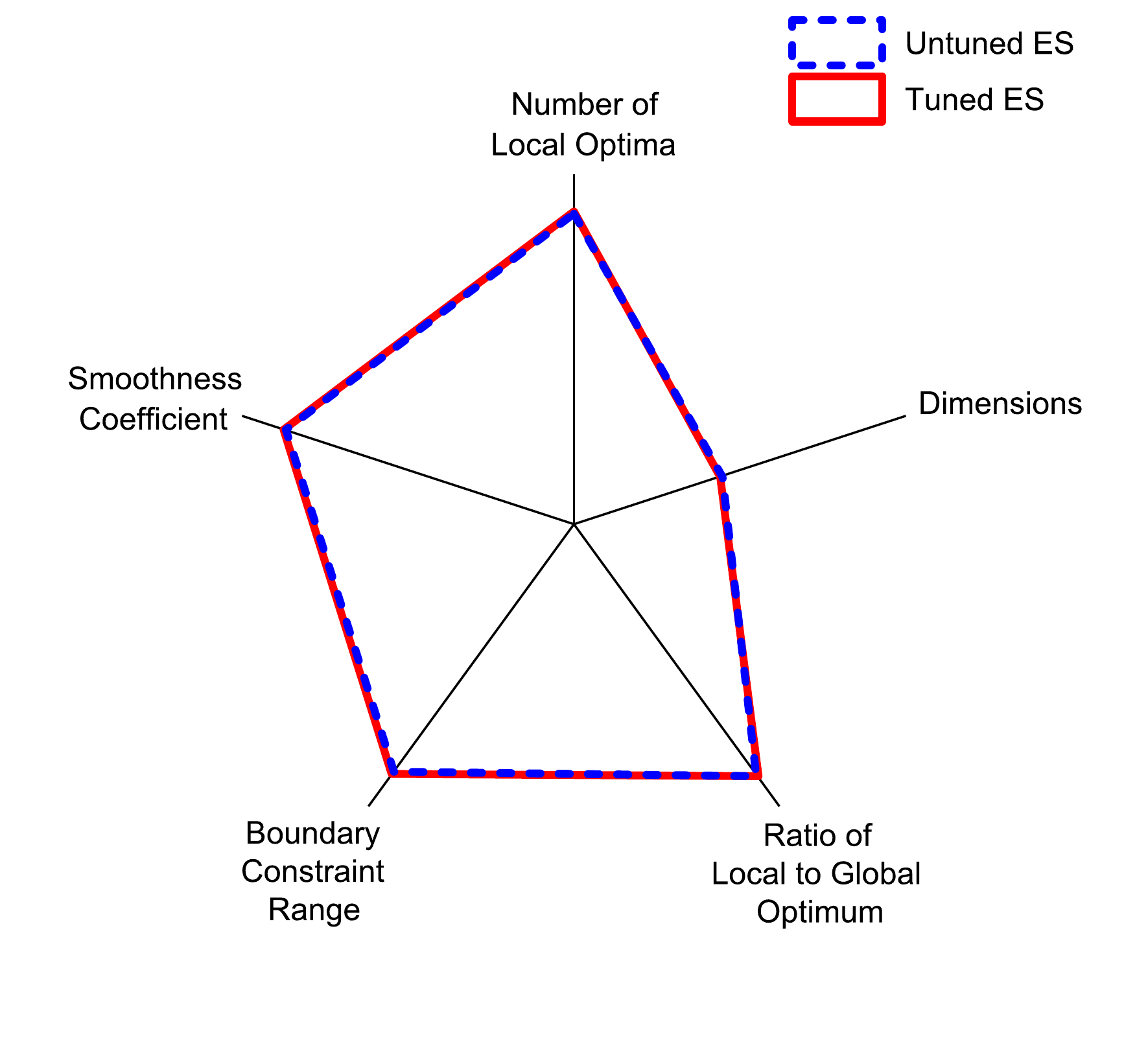}
		\caption{Summary of results for  {\bf Evolution Strategies}.}
		\label{essp}
	\end{figure}

ES has the smallest number of parameters of all the algorithms studied here (excepting the baseline algorithm, stochastic hill climbing).  The two parameters this form of ES offers are (1) {\it population size} and (2) {\it number of children}.  It is suggested \citep{selectionpressure} that altering these parameters adjusts {\it selection pressure} (that is to say, the {\it greediness} of the algorithm changes).  The parameter configurations obtained through F-Racing are varied, implying that there do exist some configurations that are more successful than others.  A range of configurations are selected across each characteristic - both in terms of different values for the two parameters, and different selection pressures when the two parameters are combined.   Results for ES are shown in Figure ~\ref{essp}. It is perhaps surprising to observe that the results of using the tuned parameters show little or no change in performance across all characteristics.  There is a small decrease in average error as the number of local optima changes, but the standard deviation is similar for both untuned and tuned, suggesting that while the average error has decreased very slightly, the ability of the algorithm to cope with increasing numbers of local optima is unchanged. For all other characteristics post-tuning, there is little change in both average error and standard deviation across characteristics values (that is to say, the algorithm is no more capable of dealing with changes in these characteristics).  This is perhaps consistent with the definition of the two parameters the algorithm offers - selection pressure can only affect the way in which ES explores local optima, and there is no control over the area that is explored around each point of interest, or any way to encourage the algorithm to rapidly explore an increasingly large search space.

We use a simple variant of ES, here, and there exist many other versions of the ES algorithm that offer a greater range of parameters (such as CMA-ES \citep{hansen2004evaluating}).  ES clearly yields its best performance with an ``out-of-the-box" parameter configuration, which means that it is quick to implement.
However, our results suggest that there is little that can be done to improve the performance of this particular variant.

\subsection{Genetic Algorithm}

\begin{figure}[t]
		\centering
		\includegraphics[width=0.4\textwidth]{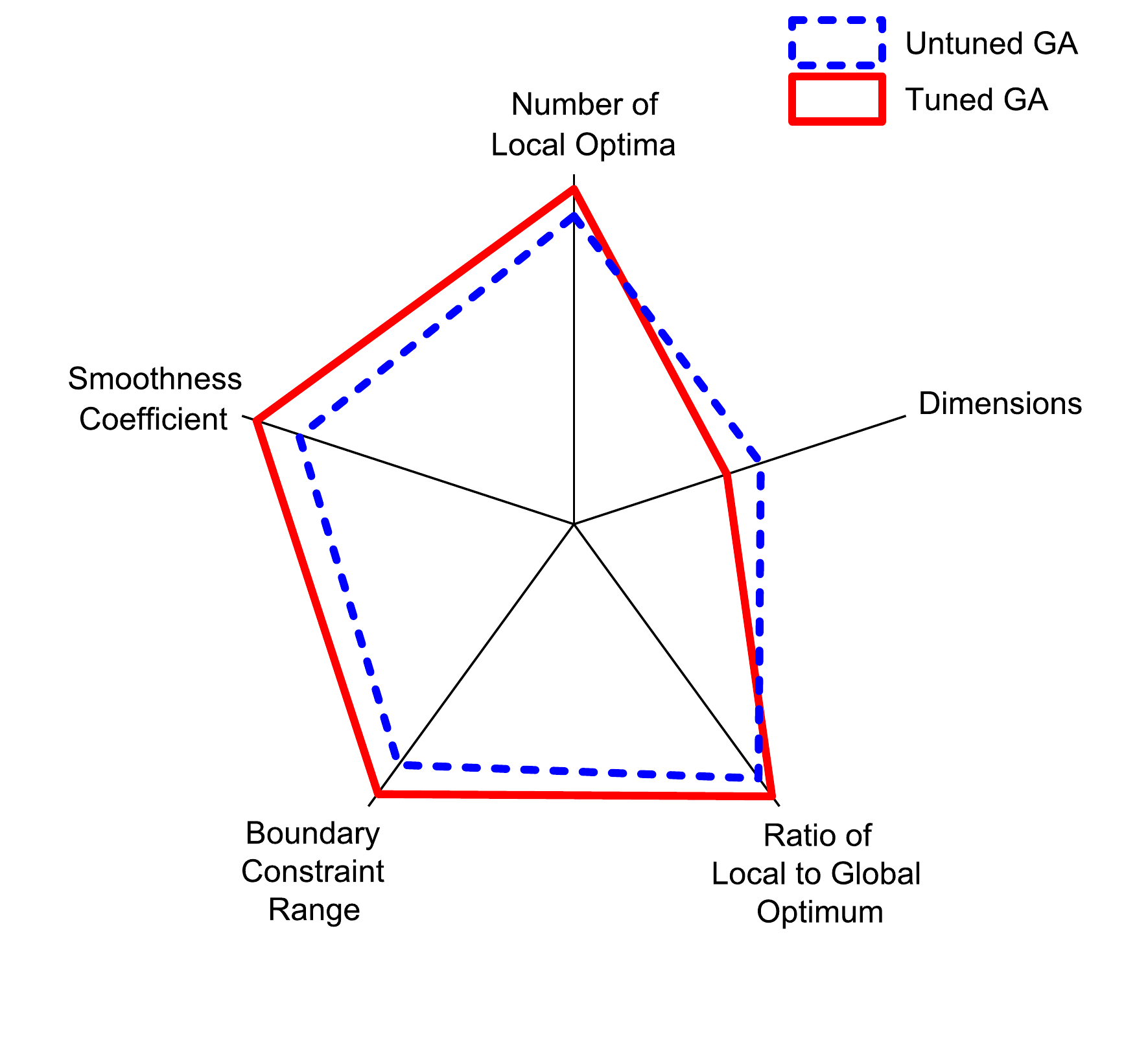}
		\caption{Summary of results for {\bf Genetic Algorithm}.}
		\label{gasp}
	\end{figure}

The performance of the GA increases post-tuning, coping significantly better with increasing numbers of local optima, increasing boundary constraint range and an increasing smoothness coefficient.  Results for the GA are shown in Figure \ref{gasp}.  The parameters of the GA are not as intuitively linked to the exploration pattern as many of the other algorithms in the study.  This particular GA offers four configurable parameters: (1) {\it population size}, (2) {\it `bits' per parameter} in the representation, (3) {\it crossover rate} and (4) {\it mutation rate}.  In experiments with a fixed number of objective function calculations, population size affects the number of generations the algorithm evaluates before terminating.  A larger number of bits in a bit string representation allows more `precise' solutions to be generated at the expense of having a representation which is less affected by mutation.  Similarly to BFOA, there are a few configurations which re-occur across different characteristics and different characteristic values.  It is probable that once a `good' configuration has been found for a GA, it is applicable to `similar' landscapes, which is consistent with the suggestion \citet{Goldberg1989} that GAs are robust problem solvers, exhibiting approximately the same performance across a wide range of problems.  

With increasing dimensionality, the GA initially shows promising results in terms of tuned performance, with a marked performance increase up to four dimensions.  The benefit from tuning rapidly declines, however, until the tuned performance is {\it worse} than that of the tuned version.  There are two possible explanations for this: the first is that the restriction on the number of objective calculations did not allow the F-Race algorithm to gather any meaningful performance data from the configurations.  The second explanation is that we did not test a wide enough range of configurations - although two of the four parameters have definite ranges (mutation and crossover rates are percentages, thus generation was bounded between zero and one), so this is unlikely. 
			
\subsection{Harmony Search}

	\begin{figure}[t]
		\centering
		\includegraphics[width=0.4\textwidth]{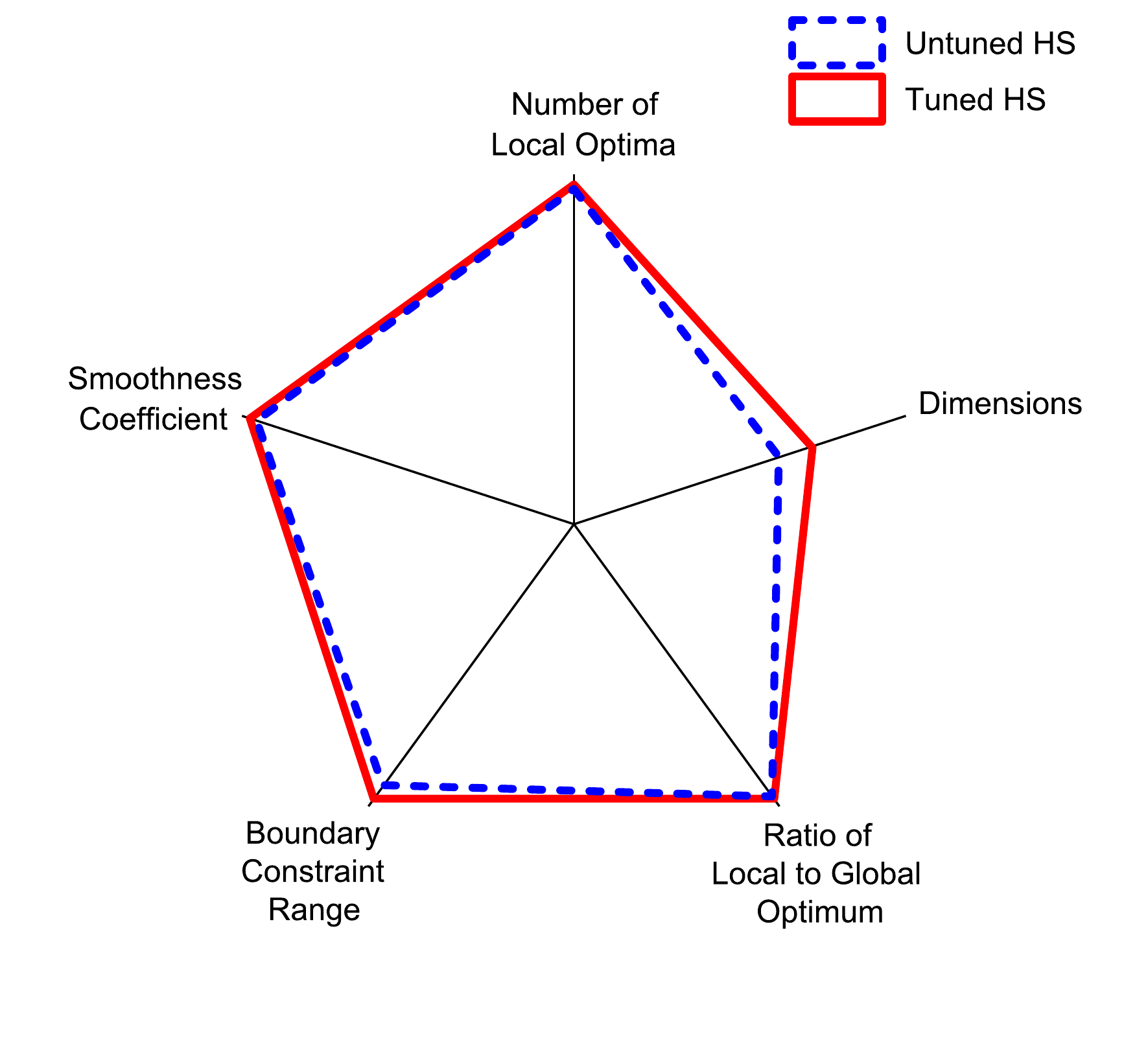}
		\caption{Summary of results for {\bf Harmony Search}.}
		\label{harmonysp}
	\end{figure}

The four parameters of HS all control different aspects of the search strategy.  {\it Memory size} dictates how many promising solutions can be stored - effectively, how many potential sites of interest are retained by the algorithm.  {\it Consideration rate} and {\it adjustment rate} control how new solutions are generated.  The consideration rate is the percentage chance that a solution based on one in memory will be generated (conversely, {\it 1-consideration rate} is the chance a random solution is generated instead).  The adjustment rate is then the percentage chance that the randomly chosen solution from memory will be adjusted.  If so, the fourth parameter, which controls the {\it maximum range} at which solutions can be adjusted, is used.  If the adjustment does not occur, the considered solution potentially occupies an additional slot in the memory - thus increasing the chance that this solution may be chosen for consideration again.  The interplay between these parameters is crucial, and it is somewhat hard to see how consideration rate and adjustment rate can directly affect the search strategy - unlike memory size and range, which are more obvious. The results for HS are shown in Figure ~\ref{harmonysp}. HS, like the BA, offers some of the lowest `out of the box' average error rates in this study.  For most characteristics, there is little room for a performance increase post-tuning.  Boundary constraint range proves to be the second-most challenging characteristic to HS pre-tuning, but post-tuning shows improved performance.  The range values in all the configurations selected by F-Racing are much smaller than those in the `out of the box' values, and this contributes significantly to the performance improvement when boundary constraint ranges are increasing.  The consideration rate also decreases almost linearly as size increases - effectively, more random solutions are used instead of relying on the `memory'.  These random solutions allow the solution pool to jump from one position in the search space to another, encouraging a wider search space, and explaining the significant improvement as boundary constraint range increases.   Dimensionality also yields an improvement in the tuned parameter performance of HS, in terms of both average error and ability to cope, as it rises.  High dimension problems (seven and above) have a much higher consideration rate than the successful configurations for lower dimensionality, suggesting that a focus on {\it exploitation} rather than {\it exploration} is beneficial to the HS when dimensionality is high.  This is the opposite case of what happens with boundary constraint range, as discussed above.

	\subsection{Particle Swarm Optimisation}

\begin{figure}[t]
		\centering
		\includegraphics[width=0.4\textwidth]{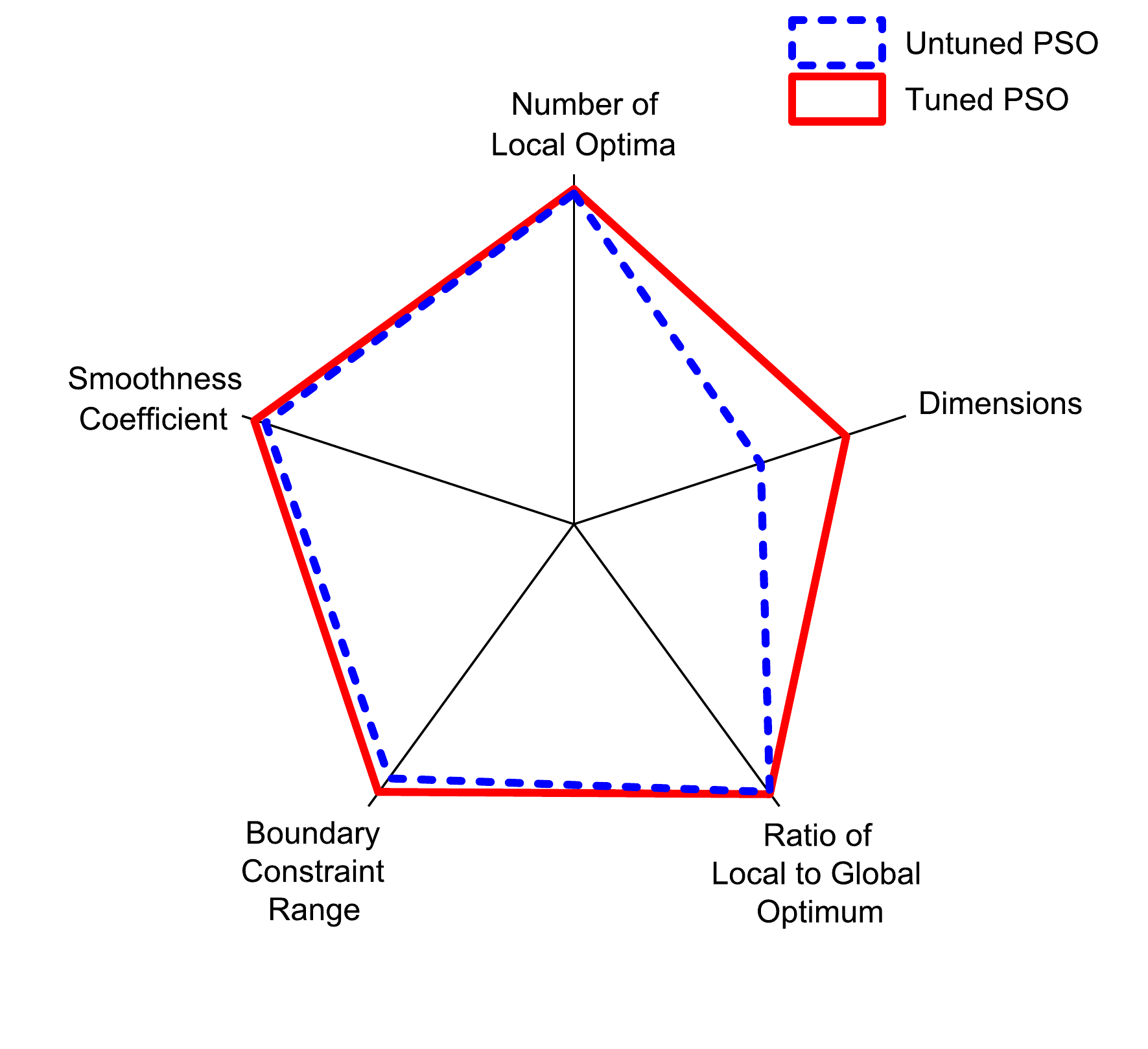}
		\caption{Summary of results for {\bf Particle Swarm Optimisation}.}
		\label{psosp}
	\end{figure}

PSO in this form has four parameters; these control the {\it population size}, the {\it maximum velocity} of a particle, the bias towards the {\it particle best solution} and the bias towards the {\it global best solution}.  With these parameters, it is possible to control the coverage of a search space (the number of particles), enforce a large search area of a small search area for each particle (the maximum velocity), and, through manipulation of the local and global best solution bias, control the capability of the algorithm to converge on a single solution or explore several areas of interest (optima avoidance). Results for PSO are shown in Figure \ref{psosp}. The parameters  used cover a broad range of search behaviours, and, as such, we would expect to see a large improvement in particle swarm performance post-tuning.  This holds true for most of our characteristics.  Results for the number of local optima, for example, show a reasonable decrease in average error as the number of local optima increases, yet the standard deviation demonstrates no change, indicating that the algorithm is no more capable of dealing with increasing numbers of local optima post-tuning.  Performance of PSO greatly improves on dimensionality post-tuning, in terms of both average error and ability to cope as it grows.  The F-Race algorithm for PSO selects the same configuration for all values of dimensionality (except for 2 dimensions), implying that there is no specific parameter that requires adjustment to cope with the increase in dimensionality, but selecting a configuration which provides good {\it exploration} allows PSO to perform well as the size of the search space increases exponentially. This trend continues across all characteristics, with F-Races often selecting the same configurations, regardless of characteristic values.  As with the other swarming algorithms, we suggest that once a good configuration has been found, it is able to deal with a wide range of problems of a similar nature, regardless of the specific characteristics.  The configurations selected are all varied in their parameters, and it is unexpected to see that there is no pattern to maximum velocity as boundary constraint range increases.  This is possibly because maximum velocity is an upper bound, and there are particles with randomly generated velocities below the maximum, so this parameter is less significant than it may initially appear.  It would perhaps be interesting to consider the effect of having a {\it minimum} velocity on the increase in boundary constraint range, although this would also severely hamper exploitation.

\subsection{Stochastic Hill-Climbing}

\begin{figure}[t]
		\centering
		\includegraphics[width=0.4\textwidth]{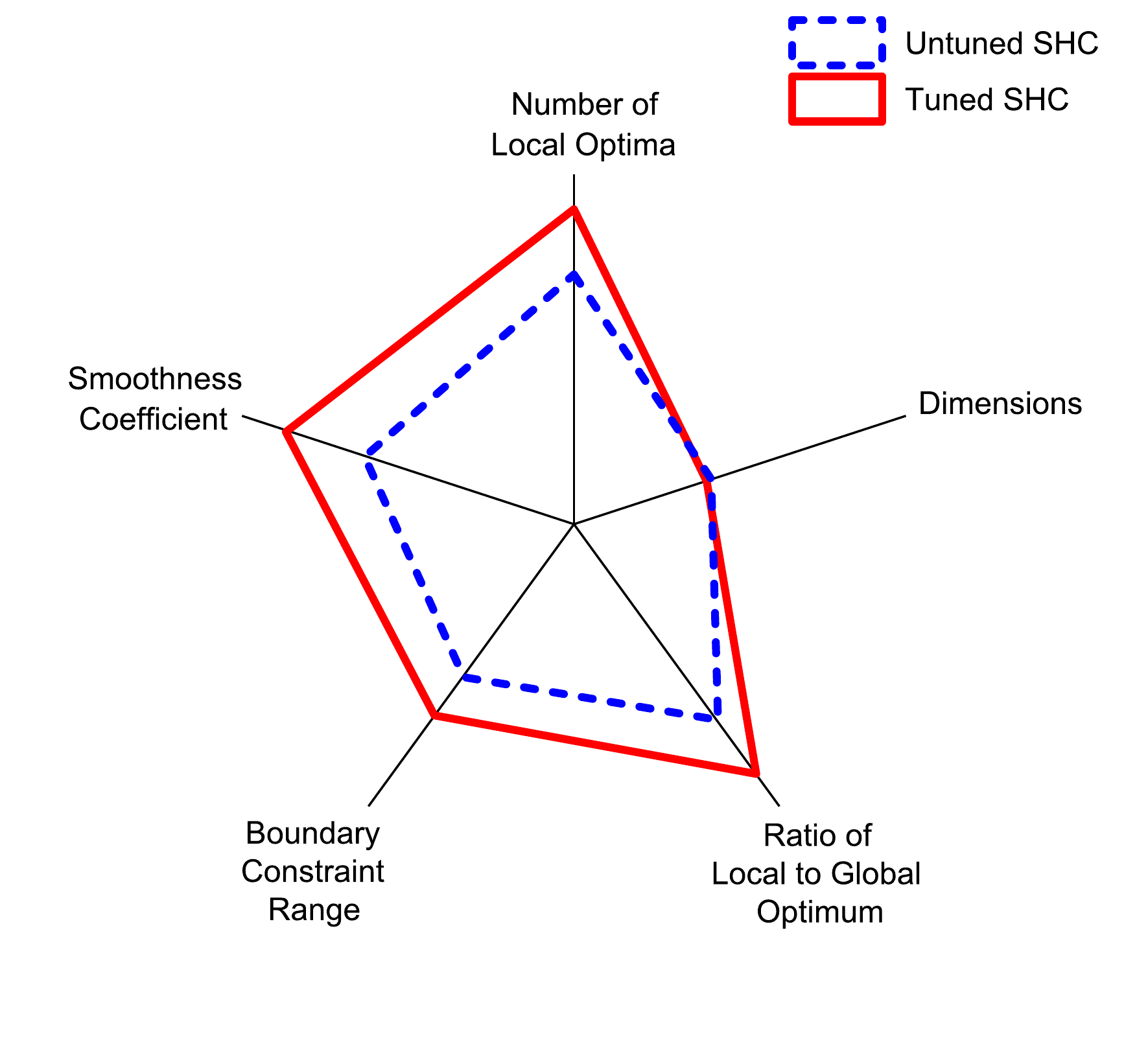}
		\caption{Summary of results for {\bf Stochastic Hill-Climbing Algorithm}.}
		\label{shcsp}
	\end{figure}

	With only a single parameter - the {\it range} at which new solutions are generated - the SHC algorithm does not offer a large amount of customisation.  This single parameter is directly linked to the search pattern and nothing else, and as there are no other parameters there is no interplay between parameters to consider.  Arguably, therefore, SHC should prove the easiest algorithm to tune. Results for SHC are shown in Figure ~\ref{shcsp}. All characteristics, barring dimensionality, show an improvement post-tuning.  As the neighbourhood size is the range at which new solutions are generated, it is unsurprising that tuning improves algorithm performance as boundary constraint ranges change.  As the number of objective function calculations is limited, despite having a larger neighbourhood size, the ability of the algorithm to effectively explore larger environments is still restricted, therefore the average error does not decrease by as much as may be expected, and the ability of the algorithm to deal with increasing search space sizes improves only slightly. SHC demonstrates a large increase in performance and a greater ability to cope with more optima (a reduced standard deviation) post-tuning.  The parameter configurations selected for the number of local optima, the ratio of local optima {\it and} the smoothness all have a neighbourhood size of around 50\% of the search space size.  We suggest that the performance improvement for all of these characteristics is actually derived from the algorithm having configured itself properly for the search space size used as a default for all other characteristics, rather than tuning itself to best perform on any specific characteristic.

\section{Conclusions and Future Work}

In this paper we have built on previous studies of the performance of nature-inspired algorithms on fitness landscapes with different characteristics.  Earlier work explored `out of the box' parameter configurations, and we futher develop this by using an automated parameter configuration methodology. This allows us to study the effect of tuning on different algorithms, contributing significantly to the debate on when and how it is beneficial to tune specific algorithms.

We observe that algorithms broadly fall into three categories: algorithms which {\it do not/sometimes/always} benefit from tuning by F-Racing.  Dimensionality often offers the most significant improvement post-tuning in algorithms, particularly those with parameters that increase the {\it breadth} of search space (swarming algorithms are significantly better here than evolutionary algorithms).  The methodology presented here is easy to implement, is computationally inexpensive, and offers considerably more information on the performance of an algorithm than using a standard set of benchmark problems.  We hope that it will offer a framework for the experimental comparison of nature-inspired algorithms, as well as a useful set of heuristics for practitioners to use in order to decide when and how to tune their methods. Future work will focus on a comparative study of tuning techniques (i.e., in addition to F-Racing), and the application of our insights to the predictive performance ranking of methods on given problems.
	
	\small
	\bibliographystyle{apalike}
	\bibliography{ecalrefs}

\begin{thebibliography}{}

\bibitem[Adenso-Diaz and Laguna, 2006]{adenso2006fine}
Adenso-Diaz, B. and Laguna, M. (2006).
\newblock Fine-tuning of algorithms using fractional experimental designs and
  local search.
\newblock {\em Operations Research}, 54(1):99--114.

\bibitem[Akay and Karaboga, 2009]{beetuning}
Akay, B. and Karaboga, D. (2009).
\newblock Parameter tuning for the artificial bee colony algorithm.
\newblock In Nguyen, N., Kowalczyk, R., and Chen, S.-M., editors, {\em
  Computational Collective Intelligence. Semantic Web, Social Networks and
  Multiagent Systems}, volume 5796 of {\em Lecture Notes in Computer Science},
  pages 608--619. Springer Berlin Heidelberg.

\bibitem[B\"{a}ck and Schwefel, 1993]{Back1993}
B\"{a}ck, T. and Schwefel, H.-P. (1993).
\newblock {An Overview of Evolutionary Algorithms for Parameter Optimization}.
\newblock {\em Evolutionary Computation}, 1(1):1--23.

\bibitem[Barr et~al., 1995]{Barr1995}
Barr, R., Golden, B., and Kelly, J. (1995).
\newblock {Designing and reporting on computational experiments with heuristic
  methods}.
\newblock {\em Journal of Heuristics}, 1:9--32.

\bibitem[Beyer and Schwefel, 2002]{Beyer2002}
Beyer, H.-g. and Schwefel, H.-p. (2002).
\newblock {Evolution strategies}.
\newblock {\em Natural Computing}, 1:3--52.

\bibitem[Birattari, 2009]{birattari2009tuning}
Birattari, M. (2009).
\newblock {\em Tuning metaheuristics: a machine learning perspective}.
\newblock Springer.

\bibitem[Birattari et~al., 2002]{birattari2002racing}
Birattari, M., St\"{u}tzle, T., Paquete, L., and Varrentrapp, K. (2002).
\newblock A racing algorithm for configuring metaheuristics.
\newblock In {\em Proceedings of the Genetic and Evolutionary Computation
  Conference}, GECCO '02, pages 11--18, San Francisco, CA, USA. Morgan Kaufmann
  Publishers Inc.

\bibitem[Birattari et~al., 2010]{birattari2010f}
Birattari, M., Yuan, Z., Balaprakash, P., and St{\"u}tzle, T. (2010).
\newblock F-race and iterated f-race: An overview.
\newblock {\em Experimental methods for the analysis of optimization
  algorithms}, pages 311--336.

\bibitem[Brabazon and O'Neill, 2006]{Brabazon2006}
Brabazon, A. and O'Neill, M. (2006).
\newblock {\em {Biologically inspired algorithms for financial modelling}}.
\newblock Springer-Verlag.

\bibitem[Brownlee, 2011]{Brownlee2011}
Brownlee, J. (2011).
\newblock {\em {Clever Algorithms: Nature-Inspired Programming Recipes}}.
\newblock Lulu.

\bibitem[Cantú-Paz, 2001]{selectionpressure}
Cantú-Paz, E. (2001).
\newblock Migration policies, selection pressure, and parallel evolutionary
  algorithms.
\newblock {\em Journal of Heuristics}, 7(4):311--334.

\bibitem[Crossley et~al., 2013]{crossley2013}
Crossley, M., Nisbet, A., and Amos, M. (2013).
\newblock Fitness landscape-based characterisation of nature-inspired
  algorithms.
\newblock In Tomassini, M., Antonioni, A., Daolio, F., and Buesser, P.,
  editors, {\em Proceedings of the 11th International Conference on Adaptive
  and Natural Computing Algorithms (ICANNGA'13), Lausanne, Switzerland, April
  4-6, 2013. Lecture Notes in Computer Science, Vol. 7824}, pages 110--119.
  Springer.

\bibitem[Eiben and Smit, 2011]{eiben2011parameter}
Eiben, A.~E. and Smit, S.~K. (2011).
\newblock Parameter tuning for configuring and analyzing evolutionary
  algorithms.
\newblock {\em Swarm and Evolutionary Computation}, 1(1):19--31.

\bibitem[Gallagher and Yuan, 2006]{Gallagher2006}
Gallagher, M. and Yuan, B. (2006).
\newblock {A general-purpose tunable landscape generator}.
\newblock {\em IEEE Transactions on Evolutionary Computation}, 10(5):590--603.

\bibitem[Geem and Kim, 2001]{Geem2001}
Geem, Z. and Kim, J. (2001).
\newblock {A new heuristic optimization algorithm: harmony search}.
\newblock {\em Simulation}, 76(2):60--68.

\bibitem[Goldberg, 1989]{Goldberg1989}
Goldberg, D.~E. (1989).
\newblock {\em {Genetic Algorithms in Search, Optimization, and Machine
  Learning}}.
\newblock Addison-Wesley.

\bibitem[Hansen and Kern, 2004]{hansen2004evaluating}
Hansen, N. and Kern, S. (2004).
\newblock Evaluating the cma evolution strategy on multimodal test functions.
\newblock In {\em Parallel Problem Solving from Nature-PPSN VIII}, pages
  282--291. Springer.

\bibitem[Hendtlass, 2009]{Hendtlass2009}
Hendtlass, T. (2009).
\newblock {Particle swarm optimisation and high dimensional problem spaces}.
\newblock In {\em 2009 IEEE Congress on Evolutionary Computation, CEC'09.},
  pages 1988--1994. IEEE.

\bibitem[Horn and Goldberg, 1994]{Horn1994}
Horn, J. and Goldberg, D. (1994).
\newblock {Genetic algorithm difficulty and the modality of fitness
  landscapes}.
\newblock In {\em Foundations of Genetic Algorithms 3}.

\bibitem[Jones and Forrest, 1995]{Jones1995}
Jones, T. and Forrest, S. (1995).
\newblock {Fitness distance correlation as a measure of problem difficulty for
  genetic algorithms}.
\newblock In {\em Proceedings of the 6th International Conference on Genetic
  Algorithms}, pages 184--192.

\bibitem[Kennedy and Eberhart, 1995]{Kennedy1995}
Kennedy, J. and Eberhart, R. (1995).
\newblock {Particle swarm optimization}.
\newblock In {\em Neural Networks, 1995. Proceedings. \ldots}, pages
  1942--1948.

\bibitem[Koster and Beney, 2007]{koster2007importance}
Koster, C. and Beney, J. (2007).
\newblock On the importance of parameter tuning in text categorization.
\newblock {\em Perspectives of Systems Informatics}, pages 270--283.

\bibitem[Kukkonen and Lampinen, 2005]{Kukkonen}
Kukkonen, S. and Lampinen, J. (2005).
\newblock {GDE3: The third Evolution Step of Generalized Differential
  Evolution}.
\newblock {\em 2005 IEEE Congress on Evolutionary Computation}, pages 443--450.

\bibitem[Leung et~al., 2003]{leung2003tuning}
Leung, F.~H., Lam, H., Ling, S., and Tam, P.~K. (2003).
\newblock Tuning of the structure and parameters of a neural network using an
  improved genetic algorithm.
\newblock {\em Neural Networks, IEEE Transactions on}, 14(1):79--88.

\bibitem[Lobo et~al., 2007]{lobo2007parameter}
Lobo, F.~G., Lima, C.~F., and Michalewicz, Z. (2007).
\newblock {\em Parameter setting in evolutionary algorithms}.
\newblock Springer Verlag.

\bibitem[Malan and Engelbrecht, 2009]{Malan2009}
Malan, K.~M. and Engelbrecht, A.~P. (2009).
\newblock {Quantifying ruggedness of continuous landscapes using entropy}.
\newblock In {\em 2009 IEEE Congress on Evolutionary Computation}, pages
  1440--1447. IEEE.

\bibitem[Maron and Moore, 1993]{maron1993hoeffding}
Maron, O. and Moore, A. (1993).
\newblock Hoeffding races: Accelerating model selection search for
  classification and function approximation.
\newblock {\em Robotics Institute}, page 263.

\bibitem[Maron and Moore, 1997]{Maron97theracing}
Maron, O. and Moore, A.~W. (1997).
\newblock The racing algorithm: Model selection for lazy learners.
\newblock {\em Artificial Intelligence Review}, 11:193--225.

\bibitem[Merz, 2000]{Freisleben2000}
Merz, P. (2000).
\newblock {Fitness landscape analysis and memetic algorithms for the quadratic
  assignment problem}.
\newblock {\em Evolutionary Computation, IEEE}, 4(4):337--352.

\bibitem[Morgan and Gallagher, 2010]{Morgan2011}
Morgan, R. and Gallagher, M. (2010).
\newblock {When does dependency modelling help? Using a randomized landscape
  generator to compare algorithms in terms of problem structure}.
\newblock In et~al Schaefer, R., editor, {\em PPSN XI}, pages 94--103.
  Springer-Verlag.

\bibitem[Nannen et~al., 2008]{nannen2008costs}
Nannen, V., Smit, S.~K., and Eiben, A.~E. (2008).
\newblock Costs and benefits of tuning parameters of evolutionary algorithms.
\newblock In {\em Parallel Problem Solving from Nature--PPSN X}, pages
  528--538. Springer.

\bibitem[Passino, 2002]{Passino2002}
Passino, K. (2002).
\newblock {Biomimicry of bacterial foraging for distributed optimization and
  control}.
\newblock {\em IEEE Control Systems Magazine}, 22(3):52--67.

\bibitem[Pham et~al., 2006]{Pham2006}
Pham, D., Ghanbarzadeh, A., and Koc, E. (2006).
\newblock {The Bees Algorithm – A Novel Tool for Complex Optimisation
  Problems}.
\newblock In Pham, D., Eldukhri, E., and Soroka, A., editors, {\em Intelligent
  Production Machines and Systems}, pages 454--459.

\bibitem[Ridge and Kudenko, 2010]{ridge2010tuning}
Ridge, E. and Kudenko, D. (2010).
\newblock Tuning an algorithm using design of experiments.
\newblock In {\em Experimental methods for the analysis of optimization
  algorithms}, pages 265--286. Springer.

\bibitem[Smit and Eiben, 2009]{smit2009comparing}
Smit, S.~K. and Eiben, A.~E. (2009).
\newblock Comparing parameter tuning methods for evolutionary algorithms.
\newblock In {\em Evolutionary Computation, 2009. CEC'09. IEEE Congress on},
  pages 399--406. IEEE.

\bibitem[Yuan and Gallagher, 2004]{yuan2004statistical}
Yuan, B. and Gallagher, M. (2004).
\newblock Statistical racing techniques for improved empirical evaluation of
  evolutionary algorithms.
\newblock In {\em Parallel Problem Solving from Nature-PPSN VIII}, pages
  172--181. Springer.

\end{thebibliography}

\end{document}